\documentclass{Interspeech2024}
\usepackage[textwidth=0.6in]{todonotes}

\interspeechcameraready

\title{Joint vs Sequential Speaker-Role Detection and Automatic Speech Recognition for Air-traffic Control}

\name[]{Alexander}{Blatt}
\name[]{Aravind}{Krishnan}
\name[]{Dietrich}{Klakow}

\address{Saarland University, Saarland Informatics Campus, Germany}

\email{ablatt@lsv.uni-saarland.de}


\begin{document}

\maketitle

\begin{abstract}
    Utilizing air-traffic control (ATC) data for downstream natural-language processing tasks requires preprocessing steps. Key steps are the transcription of the data via automatic speech recognition (ASR) and speaker diarization, respectively speaker role detection (SRD) to divide the transcripts into pilot and air-traffic controller (ATCO) transcripts. While traditional approaches take on these tasks separately, we propose a transformer-based joint ASR-SRD system that solves both tasks jointly while relying on a standard ASR architecture.  We compare this joint system against two cascaded approaches for ASR and SRD on multiple ATC datasets. Our study shows in which cases our joint system can outperform the two traditional approaches and in which cases the other architectures are preferable. We additionally evaluate how acoustic and lexical differences influence all architectures and show how to overcome them for our joint architecture.
  
\end{abstract}

    
\section{Introduction}
\label{sec:introduction}
A standard speech processing pipeline starts with a speaker diarization (SD) module, which removes the unvoiced parts of the audio and leaves speaker labeled voiced chunks. These chunks are  fed into an automatic speech recognition (ASR) system for transcription. The transcribed audio can then be further processed for example with a natural language processing (NLP) module for information extraction. Recent architectures that combine SD and ASR show however, that they can outperform this traditional pipeline \cite{Sarkar2018,Shafey2019,Xia2022,Huang2022,Cornell2023} by jointly utilizing acoustic and linguistic information during diarization.

The acoustic and linguistic information of air-traffic control (ATC) datasets however differs significantly from standard ASR and diarization datasets \cite{10389646}. ATC recordings typically have a low signal-to-noise ratio (SNR) \cite{10389646} and a strict phraseology\footnote{ATC examples: https://wiki.flightgear.org/ATC\_phraseology}, which ensures an effective communication between air-traffic  controllers (ATCOs) and  pilots. Pilot and ATCO utterances differ in the noise level as well as in the sentence structure. This can be utilized by a SD system to differentiate between the two speaker roles \texttt{ATCO} or \texttt{PILOT}, which effectively leverages it to a speaker role detection (SRD) system.

In this work, we study how a SRD system can effectively utilize the acoustic and linguistc differences between pilot and ATCO speech by analyzing the performance, respectively robustness of different ASR\&SRD architectures on multiple ATC datasets. We investigate a correlation to acoustic and linguistic properties as well as a correlation between the ASR and SRD performance. We compare three different architectures for ATC-ASR\&SRD. The first method, \texttt{SRD-ASR}, consists of an acoustic-based speaker-role detection step followed by the ASR step. The second method, \texttt{ASR-SRD}, first transcribes the audio before doing text-based SRD. Our proposed \texttt{Joint} method performs SRD and ASR simultaneously.

\section{Related work}
\label{sec:relWork}
Park et al. give good general overview of speaker diarization methods \cite{Park2022a}. Our \texttt{Joint} system is inspired by Shafey et al. which have first introduced a joint ASR\&SD system based on a recurrent neural network transducer \cite{Shafey2019}. In contrast to Shafey et al., our system performs SRD and does not require transducers, but relies on standard transformer-based ASR models \cite{Baevski2020,Babu2022} and can be trained with traditional CTC loss~\cite{graves2006connectionist}. Recent joint ASR\&SD systems require even more complex architectures than the approach of Shafey et al. \cite{Kanda2019,Xia2022,Huang2022,Cornell2023}. Our text-based SRD system is based on BERTraffic \cite{Zuluaga-Gomez2023c}, which shows a 7.7~\% improvement over a classical variational Bayesian hidden Markov model (VBx) \cite{LANDINI2022101254} based approach and is to the best of our knowledge the most recent SRD model for ATC. I a previous work, we have shown that acoustic and linguistic differences between ATC datasets negatively correlate with the performance of pretrained transformer-based ASR models \cite{10389646}. Thus, we investigate if there is a similar correlation for SRD.

\section{Datasets}
\label{sec:data}
\begin{figure}[tb]
\begin{minipage}[b]{0.43\linewidth}
  \centering
  \centerline{\includegraphics[width=1\linewidth]{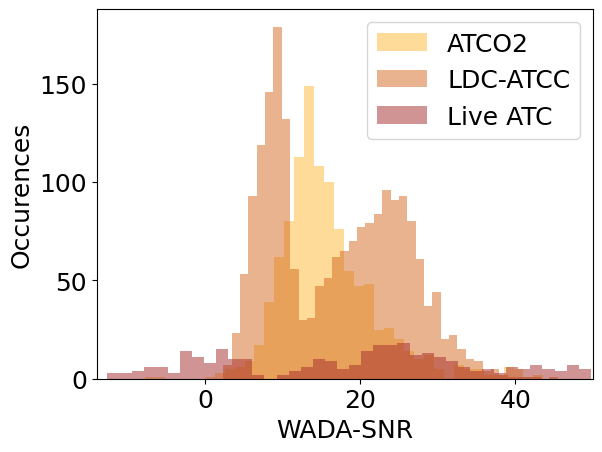}}
    \centerline{(a) Noise Distribution}\medskip
\end{minipage}
\begin{minipage}[b]{0.545\linewidth}
  \centering
  \centerline{\includegraphics[width=1\linewidth,clip]{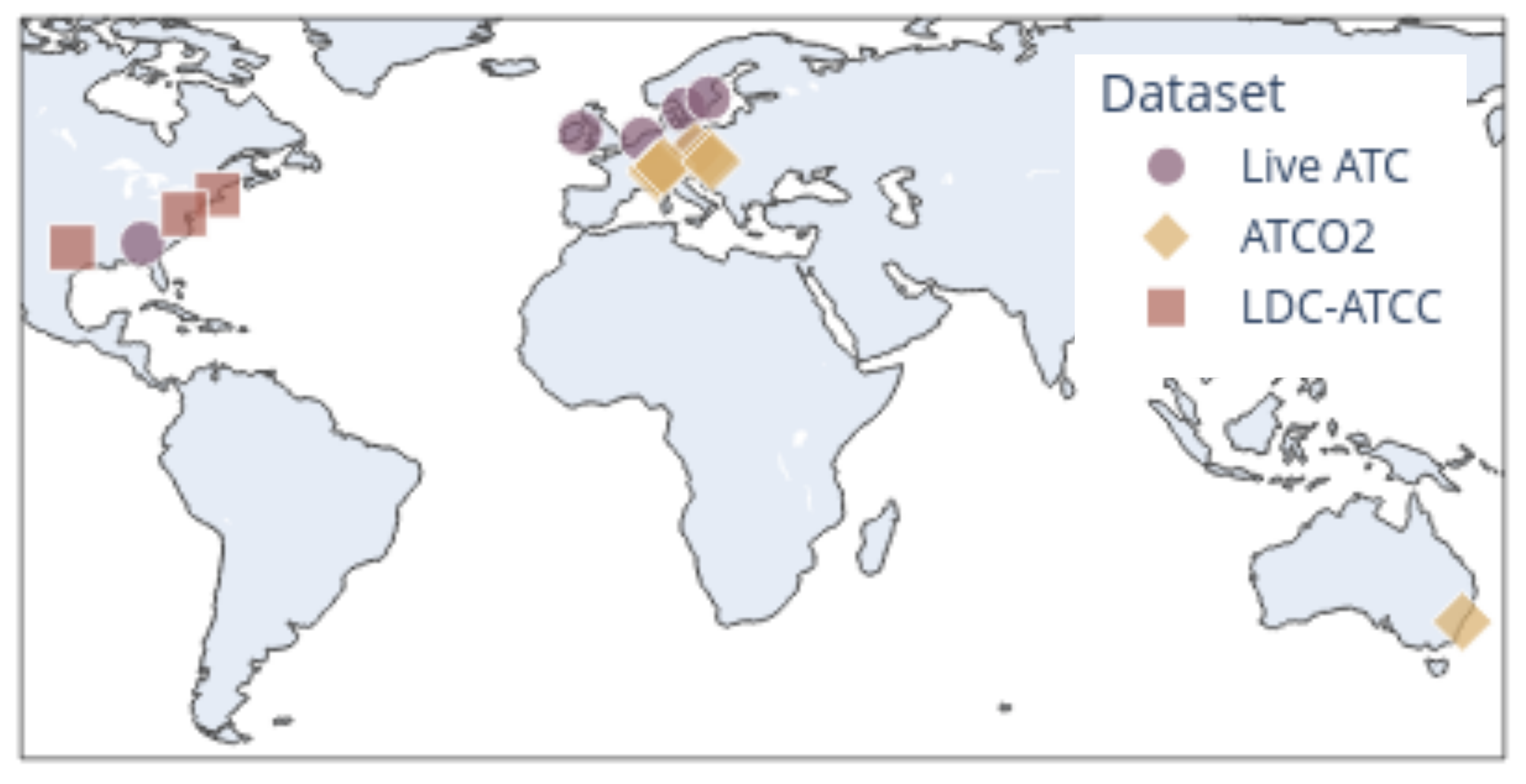}}
  \vspace{9.5pt}
\centerline{(b) Geographical Distribution}\medskip
\end{minipage}
\caption{Dataset dependent distributions}
\label{fig:distrubutions}
\end{figure}

We use the ATCO2 \cite{Zuluaga-Gomez2022b}, LiveATC \cite{Zuluaga-Gomez2020e} and the LDC-ATCC corpus \cite{Godfrey1994} for our experiments, since they all contain speaker labels, that allow to assign each speaker either to the \texttt{ATCO} or \texttt{PILOT} class. All three corpora contain ATC communication recordings. The LDC-ATCC corpus contains solely recordings from American airports, namely Dallas Fort Worth International Airport (KDFW), Logan International Airport (KBOS) and  Ronald Reagan Washington National Airport (KDCA), while the ATCO2 and LiveATC datasets contain mainly samples from European airports. They both contain samples from  Václav Havel Airport Prague (LKPR) and Zurich Airport (LSZH). The ATCO2 dataset contains additionally samples from Sion Airport (LSGS), Bratislava Airport (LZIB), Bern Airport (LSZB) and Sydney Airport (YSSY) as only non-European airport. The Live ATC dataset contains additionally samples from Stockholm Västerås Airport (ESOW), Göteborg Landvetter Airport (ESGG), Dublin Airport (EIDW), Amsterdam Airport Schiphol (EHAM) and Hartsfield–Jackson Atlanta International Airport (KATL) as only American airport. All airport locations are marked in  \autoref{fig:distrubutions}(b). The ATCO2 corpus and the LiveATC corpus were recorded during the ATCO2 project\footnote{ATCO2 project: https://www.atco2.org/}. While the ATCO2 data was recorded with VHF-receivers\footnote{Receiver guide: https://ui.atc.opensky-network.org/intro}, the LiveATC corpus, consisting of the two subcorpora LiveATC1 and LiveATC2 \cite{Zuluaga-Gomez2020e}, was recorded from the LiveATC web-page\footnote{LiveATC webpage: https://www.liveatc.net/} which broadcasts ATC conversations. The ATCO2 and LiveATC dataset audio samples are recorded with a sampling frequency of 16~kHz and 16-bit, while the LDC-ATCC data is recorded with 8~kHz and 16-bit.\par
\begin{table}[h]
\footnotesize
\centering
\caption{Number of samples for the train\textbar test\textbar val split and the mean WADA-SNR \cite{Kim2008}, mean number of speaker turns and the mean (chunked) audio  duration for each dataset.}
\label{tab:splits}
\addtolength{\tabcolsep}{-0.3em}
\begin{tabular}{lcccccc}
\toprule
Dataset  & Train & Val & Test & SNR & Turns & Duration\\
      & size & size & size & (dB) &  & (s)\\
\hline & \\[-1.5ex]
ATCO2  & 856  & 107 & 108 & 15.8&2.28 &9.4\\
LDC-ATCC & 1000 & 500 &  500 & 16.8&3.26&13.1\\
LiveATC & 413 & 42 &  41 & 18.9&2.15& 12.9\\
\bottomrule
\end{tabular}
\end{table}
Several preprocessing steps are necessary to prepare the datasets for the SRD task. To reduce the training time to a few hours per run, the original audio is chunked to samples with a target duration of 2-19 seconds. The mean chunk duration can be found in \autoref{tab:splits}. This results in 2-3 speaker turns on average as \autoref{tab:splits} shows. Samples that just contain one speaker, respectively one speaker role, are sorted out. Using the timestamps for the speaker IDs, each speaker turn is labeled with one of the two speaker roles, \texttt{ATCO} or \texttt{PILOT}. This results in transcripts, where word sequences belonging to one speaker role are tagged  with
either \texttt{ATCOTAG} or \texttt{PILOTTAG} as shown in \autoref{fig:methods}. For fine-tuning the ASR models of \texttt{SRD-ASR} and \texttt{ASR-SRD}, the tags are removed from the transcripts. 

\section{ASR\&SRD architectures}
\label{sec:mothos}
For the ASR task of all ASR\&SRD architectures, we fine-tune the Hugging Face (HF) models wav2vec 2.0\footnote{HF model: facebook/wav2vec2-base-960h} (w2v2) \cite{Baevski2020} and xlsr\footnote{HF model: jonatasgrosman/wav2vec2-large-xlsr-53-english} \cite{Babu2022} on the train split of each ATC dataset.
\begin{figure}[tb]
\begin{minipage}[b]{1\linewidth}
  \centering
  \centerline{\includegraphics[width=0.95\linewidth]{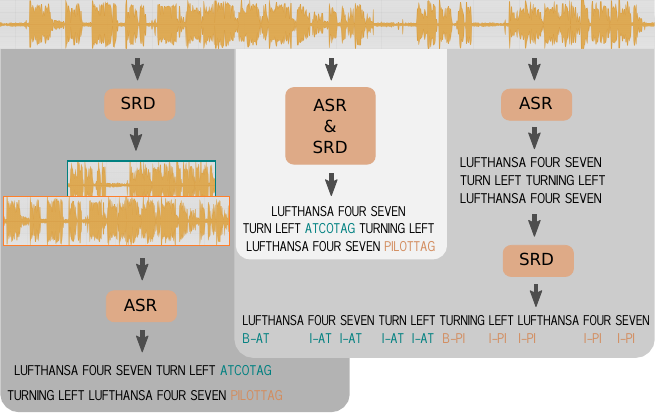}}
\end{minipage}
\caption{ASR\&SRD architectures; left: acoustic SRD followed by ASR (\texttt{SRD-ASR}); center: Joint ASR\&SRD (\texttt{Joint}); right: ASR followed by linguistic-based SRD (\texttt{ASR-SRD})}
\label{fig:methods}
\end{figure}
Each ASR\&SRD architecture is visualized in \autoref{fig:methods} and explained in the following sections.
\subsection{SRD-ASR}
For the SRD task of the \texttt{SRD-ASR} model we use the SD of Pyannote.audio 3.0\footnote{HF model: pyannote/speaker-diarization-3.0} \cite{Bredin23,Plaquet23} which combines speaker segmentation with speaker embedding based clustering for SD. The SD tool is used out-of-the-box without further fine-tuning. Just the \textit{max\_speakers} argument is set to 2, restricting diarization to two speakers. To leverage this SD to a SRD system, the extracted speakers are matched to the speaker roles by extracting the speaker embeddings of the identified speaker with the Pyannote speaker embedding extraction model\footnote{HF model: pyannote/embedding}. The classification into \texttt{PILOT} and \texttt{ATCO} is done via measuring the cosine similarity between the speaker embeddings and the cluster centers of the two speaker roles for the current training data set. The cluster centers are extracted with a nearest centroid classifier\footnote{scikit\-learn: Nearest centroid classifier} for each training data set by randomly selecting 50 samples for \texttt{PILOT} and \texttt{ATCO}. The speaker role tagged utterance chunks that are produced by the SRD system are then fed into the ASR model to generate the tagged transcripts.
\subsection{ASR-SRD}
In this approach, we train a text-based diarizer using token-level speaker labels, similar to \cite{Zuluaga-Gomez2023c}. Each word in an utterance is assigned an \texttt{ATCO} or \texttt{PILOT} tag and a binary  classifier is trained to predict the tag of each token. We encode \texttt{ATCO} and \texttt{PILOT} tags consistently across utterances to let the model learn speaker roles in addition to speaker turns. For training, ground truth transcripts from the train set are used. Testing is done on the ASR transcripts generated from the test audio. 

\subsection{Joint}
In the \texttt{Joint} approach, the ASR  models (w2v2 \& xlsr) are directly fine-tuned on the speaker role tagged transcripts instead of transcripts without tags. Since fine-tuning is done without modifying the CTC-loss function, this approach can be applied to any transformer-based ASR model with CTC loss.\begin{table*}[ht]
\footnotesize
\centering
\caption{Inter-dataset scores: WDER,PER and WER in case the models are finetuned and tested on different datasets. Mean values over three runs and two training datasets are given with the standard deviation in brackets}
\label{tab:inter}
\begin{tabular}{lcccccccccccc}
\toprule
\multirow{2}{*}{Architecture} &\multirow{2}{*} {ASR} & \multicolumn{3}{c}{ATCO2} & \multicolumn{3}{c}{LDC-ATCC} & \multicolumn{3}{c}{LiveATC} \\
\cmidrule(lr){3-5}\cmidrule(lr){6-8}\cmidrule(lr){9-11}
 & & WDER & PER &  WER  & WDER  & PER  &  WER & WDER & PER  &  WER  \\
\hline \\[-1.5ex]
SRD-ASR &  \multirow{ 3}{*}{w2v2}  &38.0 (0.5) & 32.8 (4.0) & 71.7 (4.2) &42.9 (0.1) & 37.3 (8.8) & 68.5 (2.7) & \textbf{32.0} (0.6) & 52.2 (4.2) & 82.1 (4.1)\\
ASR-SRD  & & \textbf{37.4} (0.4) & 70.5 (1.0) & \textbf{69.4} (1.5) & \textbf{40.7} (1.0) & 71.7 (1.7) & \textbf{58.7} (1.0) & 39.8 (1.1) & 71.4 (3.9) & \textbf{72.5} (2.2)\\
Joint & & 39.1 (4.2)& \textbf{19.8 (2.1)} &70.0 (1.1)& 63.4 (3.8) & \textbf{13.8 (1.9)} & 64.8 (1.0) & 38.5 (10.3) & \textbf{46.3 (4.0)} & 76.7 (3.0)\\
\hline \\[-1.5ex]
SRD-ASR & \multirow{ 3}{*}{xlsr}     &38.6 (0.5) & 31.5 (5.1) & 66.9 (5.1) &43.0 (0.3) & 35.7 (9.2) & 64.6 (9.1) & \textbf{31.3} (0.5) & 52.0 (3.9) & 76.5 (3.9)\\
ASR-SRD  & & 36.7 (0.7) & 68.4 (1.7) & \textbf{60.8} (0.8) & \textbf{39.0} (0.8) & 70.8 (2.0) & \textbf{53.3} (1.4) & 37.8 (2.5) & 69.1 (1.8) & \textbf{65.3} (1.4)\\
Joint &   & \textbf{36.6} (4.3)& \textbf{16.9} (1.7) &61.3 (1.9)& 57.9 (3.1) & \textbf{10.6 (2.8)} & 59.7 (1.1) & 33.8 (2.3) & \textbf{41.3 (1.3)} & 71.0 (2.2)\\
\bottomrule
\end{tabular}
\end{table*}\begin{table*}[b]
\footnotesize
\centering
\caption{Intra-dataset scores: WDER,PER and WER in case the models are finetuned and tested on the same dataset. Mean values over three runs are given with the standard deviation in brackets}
\label{tab:intra}
\begin{tabular}{lcccccccccccc}
\toprule
\multirow{2}{*}{Architecture} &\multirow{2}{*} {ASR} & \multicolumn{3}{c}{ATCO2} & \multicolumn{3}{c}{LDC-ATCC} & \multicolumn{3}{c}{LiveATC} \\
\cmidrule(lr){3-5}\cmidrule(lr){6-8}\cmidrule(lr){9-11}
 & & WDER & PER &  WER  & WDER  & PER  &  WER & WDER & PER  &  WER  \\
\hline \\[-1.5ex]
SRD-ASR &  \multirow{ 3}{*}{w2v2}  &27.4 (0.4) & 25.5 (7.9) & 34.8 (3.1) &27.4 (0.1) & 27.2 (5.0) & 36.2 (3.1) & 23.7 (0.4) & 51.0 (3.6) & 55.8 (4.3)\\
ASR-SRD  &  & 11.4 (0.8) & 33.5 (3.3) & 25.9 (0.4) & 12.6 (0.2) & 42.1 (1.0) & \textbf{20.2} (0.1) & 30.7 (0.5) & 80.8 (1.6) & \textbf{43.3} (0.4)\\
Joint & & \textbf{6.5} (0.4)& \textbf{4.8} (0.8) &\textbf{24.1} (0.3)& \textbf{8.0} (1.1) & \textbf{9.3} (0.5) & 27.5 (1.1) & \textbf{19.2} (7.3) &\textbf{9.3} (0.5)& 45.5 (1.8)\\
\hline\\[-1.5ex]
SRD-ASR &  \multirow{ 3}{*}{xlsr}  &27.4 (0.5) & 26.6 (9.0) & 32.0 (3.3) &27.5 (0.3) & 25.9 (4.7) & 34.4 (1.6) & 24.7 (0.4) & 50.7 (2.1) & 53.1 (0.8)\\
ASR-SRD  &  & 10.4 (0.2) & 28.6 (3.4) & \textbf{22.1} (0.6) & 12.3 (0.3) & 41.0 (0.1) & \textbf{17.6} (0.4) & 30.2 (0.2) & 81.8 (0.6) & \textbf{41.0} (0.2)\\
Joint & & \textbf{9.9} (3.0)& \textbf{4.0} (1.0) &23.1 (1.5)& \textbf{6.2} (0.3) & \textbf{2.6} (0.3) & 23.9 (1.1) & \textbf{19.1} (0.5) & \textbf{12.7} (2.9) & 43.5 (0.9)\\
\bottomrule
\end{tabular}
\end{table*}

\section{Experimental setup}
All experiments are performed on a NVIDIA V100 GPU. The ASR and the \texttt{Joint} model are trained for 2000 steps, 1000 warm-up steps, a learning rate of 4e-4 and a batch size of 4 with 8 gradient accumulation steps. We choose steps instead of epochs to ensure the same number of training steps despite different sized training sets. ASR fine-tuning takes roughly 4-5 hours with these parameters. The text-based diarizer is a BERT\footnote{HF model: https://huggingface.co/google-bert/bert-base-uncased}~\cite{DBLP:conf/naacl/DevlinCLT19} model with a binary classification head on top. It is trained with a learning rate of 2e-5, 25 warmup steps and a batch size of 16. Training is terminated by an early stopping mechanism, with a patience of 5. The models with the lowest WER (for ASR), respectively word diarization error rate (WDER) (for SRD) on the validation data are used for testing. The WDER is implemented based on Shafey at al. \cite{Shafey2019}. We additionally measure the position error rate (PER) of the speaker role tokens. This PER allows to measure if a speaker role token is placed in the correct position in the sentence independently of the speaker role. We define the PER as follows:
\begin{equation}
\text{PER}=1-\frac{ t_c}{t_p}
\end{equation}
where  $t_p$ is the number of ground truth speaker role tag positions and $t_c$ is the number of correctly placed tokens at all ground truth speaker role tag positions (class independent).\par High WERs can result in missing parts of the transcripts, which not only influences the WER but also the WDER and PER. To uncouple these, we align the target and predicted transcript with the Needleman-Wunsch algorithm \cite{NEEDLEMAN1970443} and add placeholder tokens for non-transcribed words before calculating the WDER and PER. All experiments given in the following section are repeated thrice with different seeds and the mean and standard deviation are given over those three runs if not mentioned otherwise.

\section{Results}
\subsection{Inter-and intra-dataset evaluation}
The ASR\&SRD models are tested in an inter-dataset scenario, where the train and test splits come from different datasets and an intra-dataset scenario, with the train and test splits from the same dataset. The fact that pretrained transformer-based ASR models are susceptible to inter-dataset acoustic and linguistic variabilities \cite{10389646} allows to investigate the WDER and PER scores over a wide range of WERs.
The inter-dataset scores in \autoref{tab:inter} show that the \texttt{ASR-SRD} architecture outperforms the other architectures on the ATCO2 and LDC-ATCC dataset in terms of WDER when the wave2vec 2.0  model is used. On the LiveATC dataset, the \texttt{SRD-ASR} model reaches the lowest WDER despite having the highest WER. Switching from wave2vec 2.0  to xlsr results in the lowest WDER for the \texttt{Joint} model on ATCO2. The \texttt{Joint} model also benefits the most from the model change in the other metrics. Regarding the WER, the \texttt{ASR-SRD} model outperforms the others on all datasets, while the \texttt{Joint} model has the lowest PER score on all datasets by a margin.\par This holds also true for the intra-dataset scores as shown in \autoref{tab:inter}. The \texttt{Joint} model additionally has the lowest WDER on all datasets, for both, the xlsr and the wave2vec 2.0  model. Although WER scores of the \texttt{Joint} and \texttt{ASR-SRD} architecture are close in all datasets, the \texttt{ASR-SRD} model still reaches the lowest WERs in all scenarios tested. In contrast to the \texttt{SRD-ASR} architecture, the other two ASR\&SRD models can more then half their WDER scores on the ATCO2 and LDC-ATCC dataset compared to the inter-dataset scenario. This indicates that they can utilize the fact that the lexical features don't change significantly between the training and testing scenario. This is further analyzed in the next chapter.

\subsection{Relation and causation analysis for ASR\&SRD}
\begin{figure}[t]
\begin{minipage}[b]{.32\linewidth}
  \centering
  \centerline{\includegraphics[width=\linewidth]{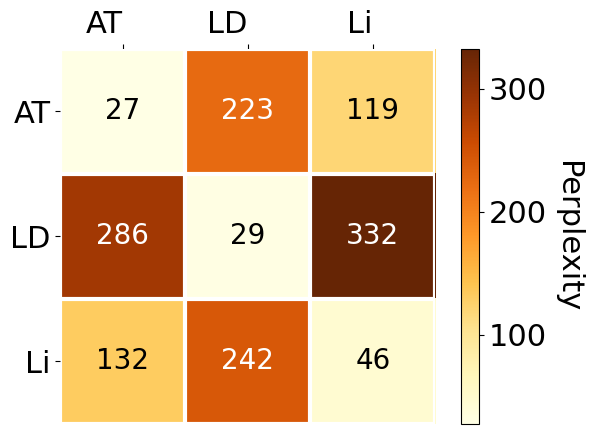}}
  \centerline{(a) Perplexity}\medskip
\end{minipage}
\hfill
\begin{minipage}[b]{0.32\linewidth}
  \centering
  \centerline{\includegraphics[width=\linewidth]{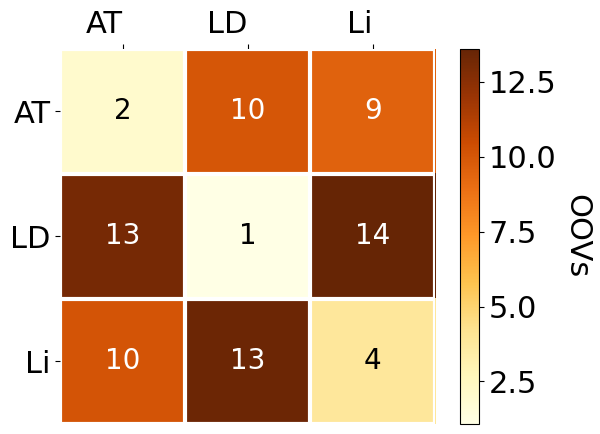}}
  \centerline{(b) OOV rate}\medskip
\end{minipage}
\begin{minipage}[b]{.32\linewidth}
  \centering
  \centerline{\includegraphics[width=\linewidth]{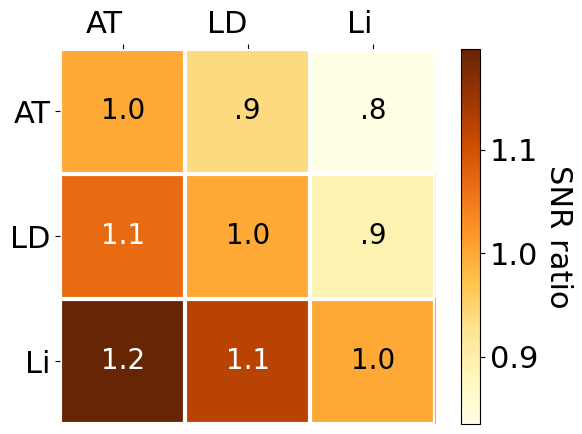}}
  \centerline{(c) SNR ratio}\medskip
\end{minipage}
\begin{minipage}[b]{.32\linewidth}
  \centering
  \centerline{\includegraphics[width=\linewidth]{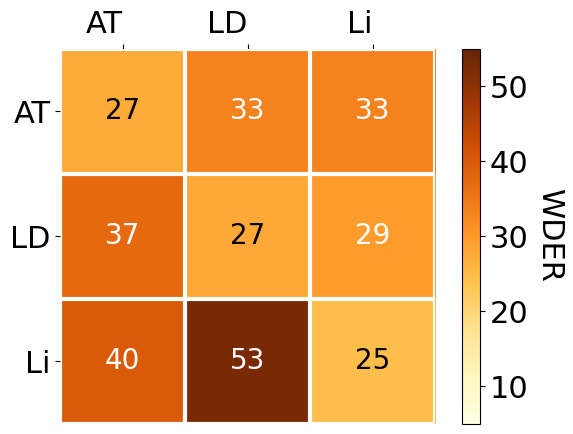}}
  \centerline{(d) WDER SRD-ASR}\medskip
\end{minipage}
\hfill
\begin{minipage}[b]{0.32\linewidth}
  \centering
  \centerline{\includegraphics[width=\linewidth]{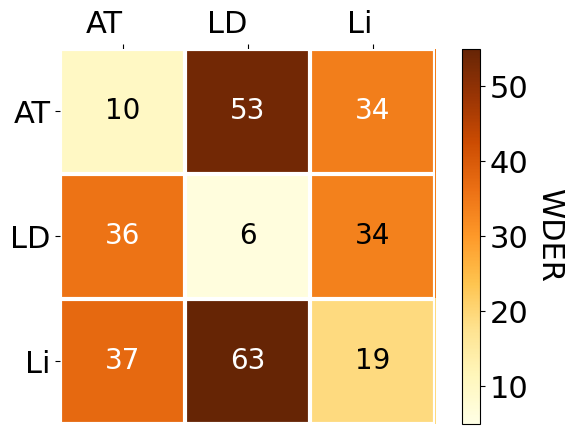}}
  \centerline{(e) WDER Joint}\medskip
\end{minipage}
\begin{minipage}[b]{.32\linewidth}
  \centering
  \centerline{\includegraphics[width=\linewidth]{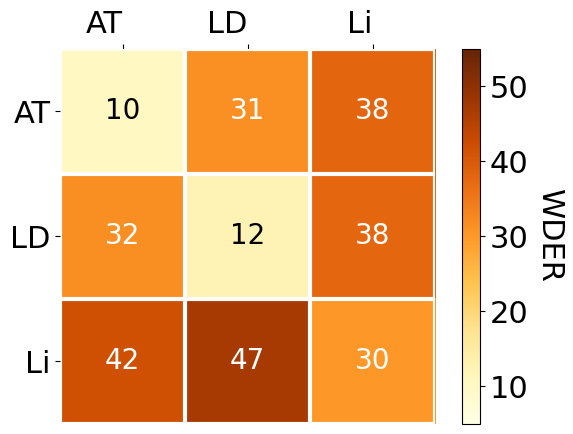}}
  \centerline{(f) WDER ASR-SRD}\medskip
\end{minipage}
\begin{minipage}[b]{.32\linewidth}
  \centering
  \centerline{\includegraphics[width=\linewidth]{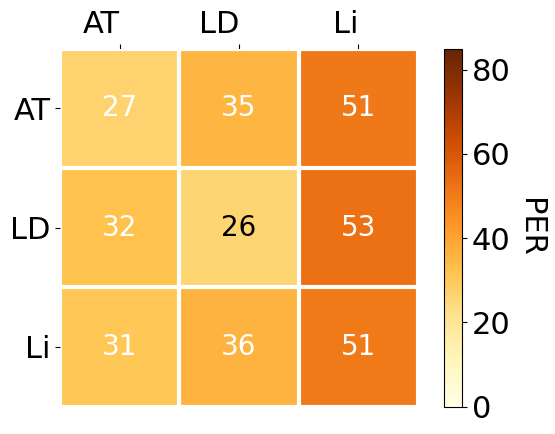}}
  \centerline{(g) PER SRD-ASR}\medskip
\end{minipage}
\hfill
\begin{minipage}[b]{0.32\linewidth}
  \centering
  \centerline{\includegraphics[width=\linewidth]{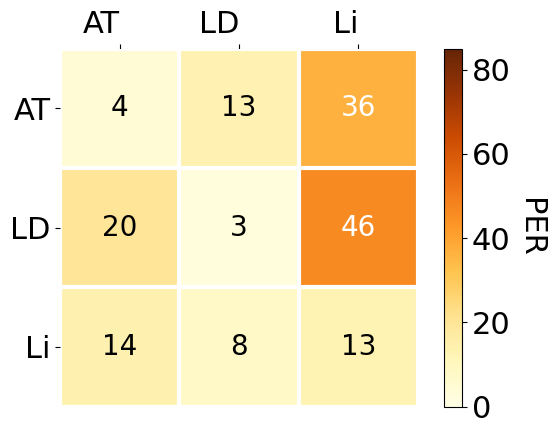}}
  \centerline{(h) PER Joint}\medskip
\end{minipage}
\begin{minipage}[b]{.32\linewidth}
  \centering
  \centerline{\includegraphics[width=\linewidth]{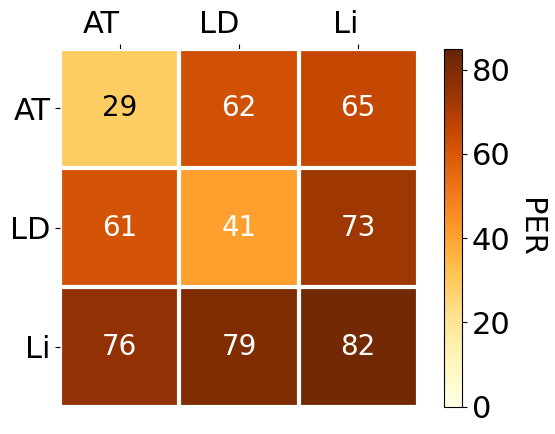}}
  \centerline{(i) PER ASR-SRD}\medskip
\end{minipage}
\begin{minipage}[b]{.32\linewidth}
  \centering
  \centerline{\includegraphics[width=\linewidth]{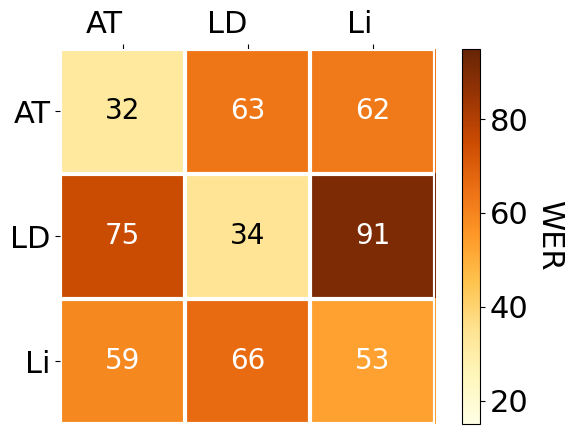}}
  \centerline{(j) WER SRD-ASR}\medskip
\end{minipage}
\hfill
\begin{minipage}[b]{0.32\linewidth}
  \centering
  \centerline{\includegraphics[width=\linewidth]{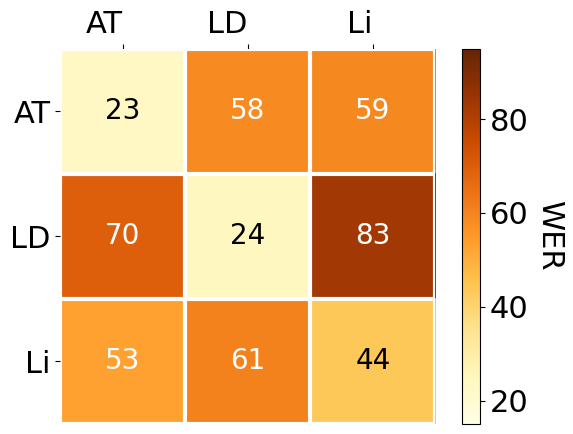}}
  \centerline{(k) WER Joint}\medskip
\end{minipage}
\begin{minipage}[b]{.32\linewidth}
  \centering
  \centerline{\includegraphics[width=\linewidth]{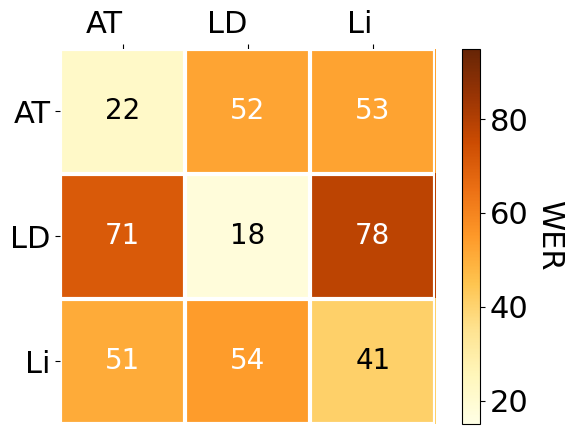}}
  \centerline{(l) WER ASR-SRD}\medskip
\end{minipage}
\caption{Confusion matrices for different metrics (a)-(l) and different ASR\&SRD methods (d)-(l) run with the xlsr model. The columns correspond to the test datasets and the rows to the training dataset. The SNR train/test ratio is calculated based on the values of \autoref{tab:splits}. The datasets are abbreviated as follows: AT: ATCO2, LD: LDC-ATCC, Li: LiveATC.}
\label{fig:confusion}
\end{figure}

To decouple/correlate the ASR and SRD performance, we analyze the confusion matrices for WDER, PER and WER in \autoref{fig:confusion}. Additional matrices for the out-of-vocabulary (OOV) rates and the perplexities allow to draw a connection to linguistic differences between the datasets. The perplexities are calculated by building a 4-gram language model (LM) on the training data and calculating the perplexity with this LM on the test data. The acoustic influences can be investigated by analyzing the SNR train/test ratio confusion matrix. The SNR values are estimated with the WADA-SNR algorithm \cite{Kim2008}. 

All three architectures have similar WER confusions matrices, the fact that the \texttt{SRD-ASR} transcribes already speaker role chunked audio files just shows in the absolute values. The PER matrices differ however significantly. The \texttt{SRD-ASR} model seems to mostly under-perform when tested on the LiveATC dataset while the \texttt{ASR-SRD} model produces high PERs when trained on the LiveATC dataset.
The \texttt{Joint} model shows a balanced performance except for the case when trained on LDC-ATCC and tested on LiveATC. This corresponds with the perplexity and OOV rate matrices, which also show a high value for this pairing.\par
The WDER matrices show that the \texttt{SRD-ASR} architecture has the most balanced performance while only producing high WDERs on the liveATC - LDC-ATCC pairing, which is also the case for the other architectures. This could be due to the fact that this pairing shows also a high perplexity, OOV rate and SNR ratio. The confusion matrix on the \texttt{Joint} model highlights the performance gap between the inter and intra-dataset scenario. The WDER, WER and PER matrices of the \texttt{ASR-SRD} value show a high similarity, indicating a correlation between the three measures. Overall, the perplexity and OOV rates seem to have a higher influence on the ASR\&SRD metrics than the SNR ratio. But it should be noted, that the WADA-SNR values of the datasets are quite similar as \autoref{tab:splits} shows. However, the distribution of the SNR values are quite different as \autoref{fig:distrubutions} (a) indicates. An additional noise analysis is therefore necessary to draw noise-related conclusions.

\begin{figure}[t]
  \centering
  \includegraphics[width=0.59\linewidth]{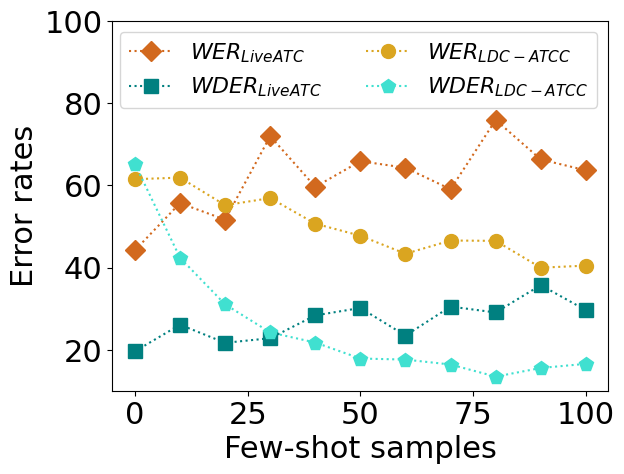}
\caption{Few-shot learning on LDC-ATCC of a \texttt{Joint}-xlsr model finetuned previously on Live ATC data. All experiments are just conducted once.}
\label{fig:fewShot}
\end{figure}
\subsection{Few-shot learning}
The difference between the inter- vs intra-dataset WDER scores for the \texttt{Joint} architecture is quite large as shown above. To ameliorate this with domain familiarization, we resort to few-shot training. As an example case, we use the LDC-ATCC data to further train a \texttt{Joint}-xlsr model finetuned on Live ATC data. \autoref{fig:fewShot} shows that the WDER on LDC-ATCC drops to 42\% by just using 10 samples from the LDC-ATCC data for fine-tuning. This is already the level that the other architecture reach on this dataset. By using 50 samples, the WDER is already under 20\%. There is however a noticeable increase in the WDER/WER on the Live ATC dataset. At 25 samples, the model shows a balanced inter-and intra-dataset performance. This shows that adaptation to the cross-dataset scenario is possible by using few-shot training. 


\section{Conclusion} 
Recently proposed joint diarization and ASR models outperform the traditional sequential approaches. The air-traffic control (ATC) domain differs however acoustically and linguistically from standard diarization and ASR datasets. In ATC, identifying the speaker role, pilot or air-traffic controller, is often more important than identifying the speaker. We have therefore proposed a joint speaker-role detection (SRD) and ASR system for ATC (\texttt{Joint}). This system purely relies on a transformer based ASR models. We have compared this architecture against two traditional cascaded approaches, that either first perform ASR, then text based SRD (\texttt{ASR-SRD}), or first acoustic based SRD and then ASR (\texttt{SRD-ASR}). Our system clearly outperforms the other systems in the intra-dataset scenario in terms of the word diarization error rate
(WDER). The position error rate (PER) scores are lower in all scenarios. We can show that the WDER scores of the (\texttt{Joint}) and (\texttt{ASR-SRD}) systems scale with a better ASR performance, while the (\texttt{Joint}) models seems to benefit more from a potent ASR model. Few-shot training results indicate that the inter-dataset scores of the \texttt{Joint} model can be significantly improved with just 25 samples. The \texttt{ASR-SRD} architecture shows a more balanced performance between the intra- and inter-dataset scenario, while the \texttt{SRD-ASR} approach only seems to be superior if there is a high WER scenario. These insights allow to pick the correct architecture for an individual ASR\&SRD task.

\bibliographystyle{IEEEtran}
\bibliography{mybib}

\end{document}